\def\year{2018}\relax
\documentclass[letterpaper]{article} 
\usepackage{aaai18}  
\usepackage{times}  
\usepackage{helvet}  
\usepackage{courier}  
\usepackage{url}  
\usepackage{graphicx}  

\usepackage{subcaption}
\usepackage[utf8]{inputenc} 
\usepackage[T1]{fontenc}    
\usepackage{booktabs}       
\usepackage{amsfonts}       
\usepackage{nicefrac}       
\usepackage{microtype}      
\usepackage{amsmath}
\usepackage{algorithm}
\usepackage[noend]{algpseudocode}
\usepackage{lipsum}

\begin{document}
%

\title{Adaptive Graph Convolutional Neural Networks}
\author{Ruoyu Li, Sheng Wang, Feiyun Zhu, Junzhou Huang\thanks{Corresponding author: J. Huang, jzhuang@uta.edu}\\
	The University of Texas at Arlington, Arlington, TX 76019, USA\\
	Tencent AI Lab, Shenzhen, 518057, China}

\maketitle
\begin{abstract}
	Graph Convolutional Neural Networks (Graph CNNs) are generalizations of classical CNNs to handle graph data such as molecular data, point could and social networks. Current filters in graph CNNs are built for fixed and shared graph structure. However, for most real data, the graph structures varies in both size and connectivity. The paper proposes a generalized and flexible graph CNN taking data of arbitrary graph structure as input. In that way a task-driven adaptive graph is learned for each graph data while training. To efficiently learn the graph, a distance metric learning is proposed. Extensive experiments on nine graph-structured datasets have demonstrated the superior performance improvement on both convergence speed and predictive accuracy.
\end{abstract}

\section{Introduction}
\label{sec:introduction}

Although the Convolutional Neural Networks (CNNs) have been proven supremely successful on a wide range of machine learning problems \cite{hinton2012deep,dundar2015simplicity}, they generally require inputs to be tensors. For instance, images and videos are modeled as 2-D and 3-D tensor separately. However, in many real problems, the data are on irregular grid or more generally in non-Euclidean domains, e.g. chemical molecules, point cloud and social networks. Instead of regularly shaped tensors, those data are better to be structured as graph, which is capable of handling varying neighborhood vertex connectivity as well as non-Euclidean metric. Under the circumstances, the stationarity and the compositionality, which allow kernel-based convolutions on grid, are no longer satisfied. Therefore, it is necessary to reformulate the convolution operator on graph structured data.

\begin{figure}[h!]
	\centering
	\includegraphics[width=0.2\textheight]{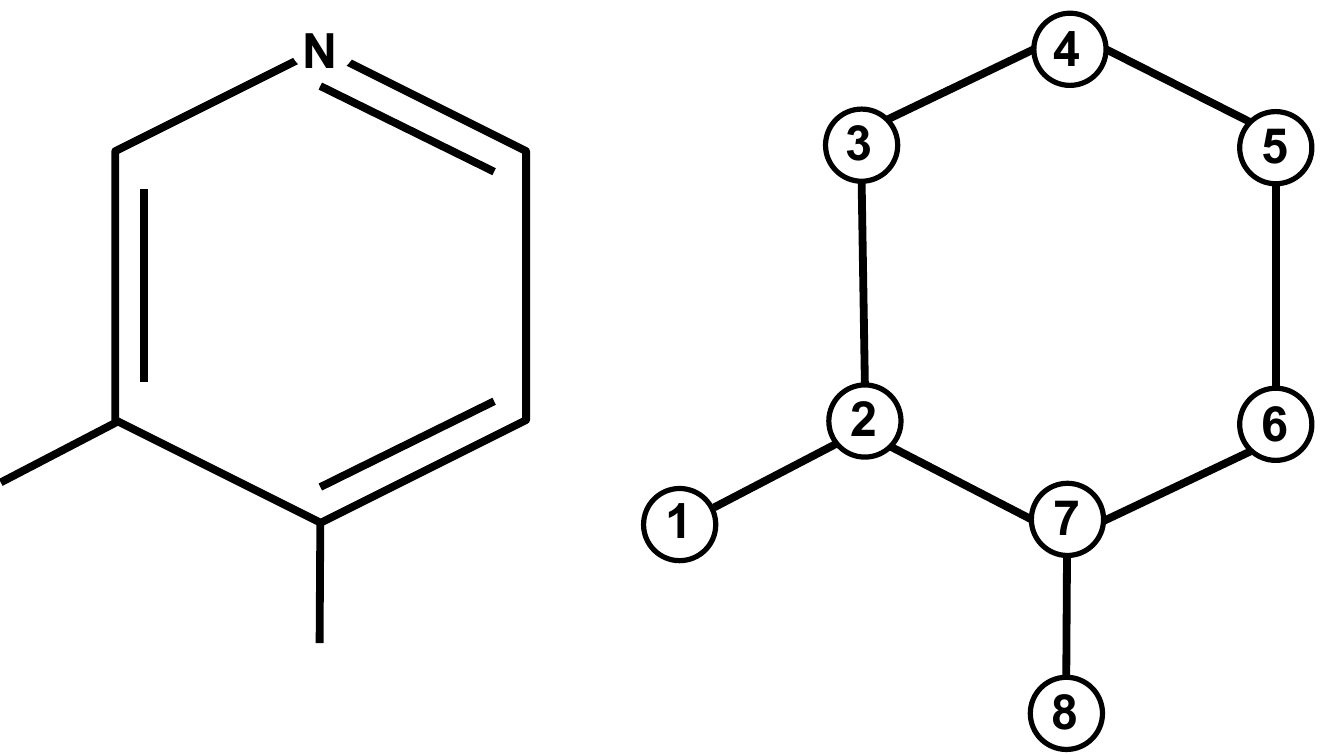}
	\caption{Example of graph-structured data: organic compound 3,4-Lutidine (C7H9N) and its graph structure.} \label{fig:example_mol}
\end{figure}

However, a feasible extension of CNNs from regular grid to irregular graph is not straightforward. For simplicity of constructing convolution kernel, the early graph CNNs usually assume that data is still low-dimensional \cite{bruna2013spectral,henaff2015deep}. Because the convolver handled nodes separately according to node degree. And their convolution kernel is over-localized and infeasible to learn hierarchical representations from complex graphs with unpredictable and flexible node connectivity, e.g molecules and social networks.

In some cases, e.g classification of point cloud, the topological structure of graph is more informative than vertex feature. Unfortunately, the existing graph convolution can not thoroughly exploit the geometric property on graph due to the difficulty of designing a parameterized spatial kernel matches a varying number of neighbors \cite{shuman2013emerging}. Besides, considering the flexibility of graph and the scale of parameter, learning a customized topology-preserving spatial kernel for every unique graph is impractical. 

Beyond spatial convolution on restricted graphs, spectral networks, based on graph Fourier transform, offer an elastic kernel \cite{defferrard2016convolutional}. Inherited from classical CNNs, a shared kernel among samples is still assumed. Consequently, to ensure the unified dimensionality of layer output, the inputs have to be resized, which is also a constraint of classical CNNs. However, this kind of preprocessing on graph data may destroy the completeness of graph-oriented information. For instance, the coarsening of molecule is hard to be justified chemically, and it is likely that the coarsened graph has lost the key sub-structures that differentiate the molecule from others. In Figure. \ref{fig:example_mol}, removing any Carbon atom from the graph breaks the Benzene ring.
It would be much better if the graph CNNs could accept original data samples of diverse graph structures.

\begin{figure*}[ht!]
	\centering
	\includegraphics[width=0.5\textheight]{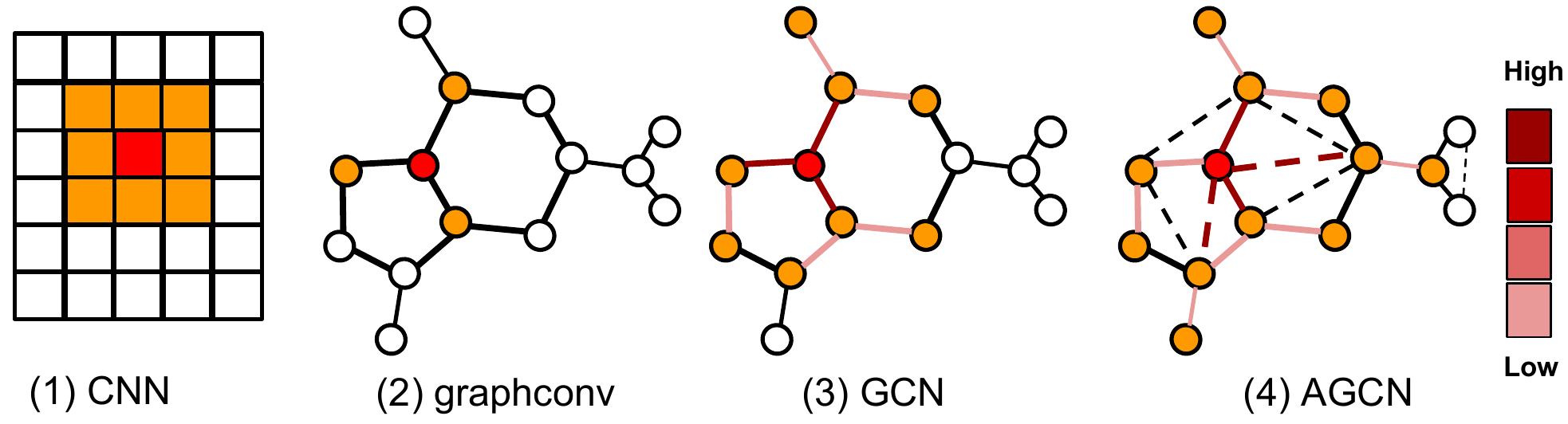}
	\caption{Convolution kernel comparison. Red point: centre of kernel. Orange points: coverage of kernel. (1) $3\times 3$ kernel of classical CNN on 2-D grid; (2) graphconv/neural fingerprint, strictly localized kernel; (3) GCN, $K$-localized kernel merely on the shared graph; (4) AGCN, $K$-localized kernel on adaptive graph (individual graph + learned residual graph). Edges from learned residual graph Laplacian are dash lines. Color of edge indicates the weights in spectral kernels. Levels of value as color bar. }\label{fig:kernels}
\end{figure*}

Lastly, the data we feed to graph CNNs either have an intrinsic graph structure or we can construct one by clustering. At previous graph CNNs, the initial graph structure will be fixed during the training process \cite{bruna2013spectral}. But, it is hard to evaluate if the graphs constructed by unsupervised clustering (or from domain knowledge) are optimal for supervised learning tasks. Although the supervised graph construction with fully connected networks has been proposed \cite{henaff2015deep}, their dense training weights restrict the model to small graphs. Furthermore, the graph structures learned from a separate network are not guaranteed to best serve the graph convolutions. 

The bottlenecks of current graph CNNs include:
\begin{itemize}
	\item restrict graph degree;
	\item require identical graph structure shared among inputs;
	\item fixed graph constructed without training;
	\item incapability of learning from topological structure.
\end{itemize}

In the paper, we propose a novel spectral graph convolution network that feed on original data of diverse graph structures. e.g the organic molecules that consist of a different number of benzene rings. To allow that, instead of shared spectral kernel, we give each individual sample in batch a customized graph Laplacian that objectively describes its unique topology. A customized graph Laplacian will lead to a customized spectral filter that combines neighbor features according to its unique graph topology.

It is interesting to question what exact graph best serves a supervised learning task. For example, the chemical bonds naturally build a graph for a compound. However, it is never guaranteed that the convolver that works on intrinsic graph has extracted all meaningful features. Therefore, we train a so-called residual graph to discover the \emph{residual} sub-structures that the intrinsic graph never includes. Moreover, to ensure that the residual graph is the best supplement for particular task, we design a scheme to learn the residual graph during training the rest of graph CNN.

Direct learning of graph Laplacian costs $\mathcal{O}(N^2)$ complexity for a $\mathbb{R}^{N\times d}$ graph of $N$ vertices. Allowing unique graph topology preserved in $M$ training samples means learning $M$ unique graph Laplacian, which is highly costly. If harnessing a supervised metric learning with \emph{Mahalanobis} distance, we could reduce the parameter number to $\mathcal{O}(d^2)$ or even $\mathcal{O}(d)$, assuming metric parameters are shared across samples. As a consequence, the learning complexity becomes independent of graph size $N$. 
In classical CNNs, back-propagation generally updates kernel weights to adjust the relationship between neighboring nodes at each feature dimension individually. Then it sums up signals from all filters to construct hidden-layer activations. To grant graph CNNs a similar capability, we propose a re-parameterization on the feature domain with additional transform weights and bias. Finally, the total $\mathcal{O}(d^2)$ training parameters in the convolution layer consist of two parts: distance metric, and the vertex feature transform and bias. Given the trained metric and transformed feature space, the updated residual graph is able to be constructed.  

In experiments, we explore the proposed spectral convolution network on multiple graph-structured datasets including chemical molecules and point cloud generated by LIDAR. The innovations of our graph CNN are summarized as below:
\begin{enumerate}
	\item \textbf{Construct unique graph Laplacian}. Construct and learn \emph{unique} residual Laplacian matrix for each individual sample in batch, and the learned residual graph Laplacian will be added onto the initial (clustered or intrinsic) one.
	\item \textbf{Learn distance metric for graph update}. Through learning the optimal distance metric parameters shared among the data, the topological structures of graph are updated along with the training of prediction network. The learning complexity is cheap as $\mathcal{O}(d^2)$, independent of input size.
	\item \textbf{Feature embedding in convolution}. Transforming of vertex features is done before convolution connecting both intra- and inter-vertex features on graph.
	\item \textbf{Accept flexible graph inputs}. Because of 1 and 2, the proposed network can be fed on data of different graph structure and size, unlocking restrictions on graph degree. 
\end{enumerate}

\section{Related Work}
\label{sec:related}

\subsection{Spectral Graph Convolution} 
\label{subsec:SGC}

The first trial of formulating an analogy of CNN on graph was accomplished by \cite{bruna2013spectral}. Particularly, the spatial convolution summed up the features from neighborhood defined by graph adjacency matrix $A_k$. The finite-size kernel is nonparametric but over-localized. The convolution layer was reduced to an analog of fully connected layer with sparse transform matrix given by $A_k$. Spatial convolution has intrinsic difficulty of matching varying local neighborhoods, so there is no unified definition of spatial filtering on graph without strict restriction on graph topology.  Spectral graph theory \cite{chung1997spectral} makes it possible to construct convolution kernel on spectrum domain, and the spatial locality is supported by the smoothness of spectrum multipliers. The baseline approach of the paper is built upon [Eq(3), \cite{defferrard2016convolutional}] that extended the one-hop local kernel to the one that brought at most $K$-hop connectivity. According to graph Fourier transform, if $U$ is the set of graph Fourier basis of $L$,
\begin{equation}
x_{k+1} = \sigma\big( g_{\theta}(L^K) x_{k} \big) = \sigma\big( U g_{\theta}(\Lambda^K) U^T x_{k} \big). \label{eq:spectral_layer}
\end{equation}
$diag(\Lambda)$ is the $\mathcal{O}(N)$ frequency components of $L$.  
\cite{defferrard2016convolutional} also utilized Chebyshev polynomials and its approximate evaluation scheme to reduce the computational cost and achieve localized filtering. \cite{kipf2016semi} showed a first-order approximation to the Chebyshev polynomials as the graph filter spectrum, which requires much less training parameters. Even though, \cite{de2016dynamic,simonovsky2017dynamic,looks2017deep} have started to construct customized graphs with more emphasis on topological structure, or even unlock the constraint on input graph dimensionality, designing a more flexible graph CNN is still an open question.

\subsection{Neural Networks on  Molecular Graph}
\label{subsec:NNM}

The predictions on checmical property of organic molecule were usually handled by handcrafted features and feature embedding \cite{mayr2016deeptox,weiss2009spectral}. Since molecules are naturally modeled as graph, \cite{duvenaud2015convolutional,wallach2015atomnet,wu2017moleculenet} have made several successful trials of constructing neural networks on raw molecules for learning representations. However, due to the constraints of spatial convolution, their networks failed to make full use of the atom-connectivities, which are more informative than the few bond features. More recent explorations on progressive network, multi-task learning and low-shot or one-shot learning have been accomplished \cite{altae2016low,gomes2017atomic}. So far, the state-of-the-art network on molecules \cite{wallach2015atomnet,duvenaud2015convolutional} still use non-parameterized spatial kernel that can not fully exploit spatial information. Besides, the topological structures can be rich sources of discriminative features.

\section{Method} 
\label{sec:method}

\subsection{SGC-LL Layer}
In order to make the spectral convolution kernel truly feasible with the diverse graph topology of data, we parameterize the distance metrics, so that the graph Laplacian itself becomes trainable.  With the trained metrics, we dynamically construct unique graph for input samples of different shape and size. The new layer conducts convolution with $K$-localized spectral filter constructed on adaptive graph. In the meanwhile,  the graph topological structures of samples get updated minimizing training losses. The new Spectral Graph Convolution layer with graph Laplacian Learning is named as SGC-LL. In this section, we introduce the innovations of SGC-LL layer.

\subsubsection{Learning Graph Laplacian}
\label{subsec:learning_laplacian}

Given graph $\mathcal{G}=(V, E)$  and its adjacency matrix $A$ and degree matrix $D$, the normalized graph Laplacian matrix $L$ is obtained by :
\begin{equation}
L = I - D^{-1/2} A D^{-1/2} \label{eq:laplacian}.
\end{equation}
Obviously, $L$ determines both the node-wise connectivity and the degree of vertices. Knowing matrix $L$ means knowing the topological structure of graph $\mathcal{G}$. Because $L$ is a symmetric positive definite matrix, its eigendecomposition gives a complete set of eigenvectors $U$ formed by $\{u_s\}_{s=0}^{N-1}$, $N$ is the number of vertices. Use $U$ as graph Fourier basis, graph Laplacian is diagonalized as $L = U \Lambda U^T$. Similar to Fourier transform on Euclidean spaces, graph Fourier transform, defined as $\hat{x}=U^T x$, converts graph signal $x$ (primarily vertex feature) to spectral domain spanned by basis $U$. Because the spectral representation of graph topology is $\Lambda$, the spectral filter $g_{\theta}(\Lambda)$ indeed generates customized convolution kernel on graph in space. \cite{chung1997spectral} tell us that a spectrum formed by smooth frequency components results in localized spatial kernel. The main contribution of \cite{defferrard2016convolutional} is to formulate $g_{\theta}(\Lambda)$ as a polynomial:
\begin{equation}
g_{\theta}(\Lambda) = \sum_{k=0}^{K-1} \theta_k \Lambda^k \label{eq:kthspectralkernel},
\end{equation}
which brings us an $K$-localized kernel that allows any pair of vertices with shortest path distance $d_{\mathcal{G}}<K$ to squeeze in. Also, the far-away connectivity means less similarity and will be assigned less importance controlled by $\theta_k$. Polynomial filter smoothen the spectrum, while parameterization by $\theta_k$ also forces a circular distribution of weights in resulted kernel from central vertex to farthest $K$-hop vertices. This restricts the flexibility of kernel. What's more important is that the similarity between two vertices is essentially determined by the selected distance metrics and the feature domain. For data deployed in non-Euclidean domain, the Euclidean distance is no longer guaranteed to be the optimal metrics for measuring similarity. Therefore, it is possible that the similarity between connected nodes is lower than those disconnected because the graphs are suboptimal. And there are two possible reasons:
\begin{itemize}
\item The graphs were constructed in raw feature domain before feature extraction and transform.
\item The graph topology is intrinsic, and it merely represents physical connections, e.g the chemical bonds in molecule.
\end{itemize}

To unlock the restrictions, we propose a new spectral filter that parameterizes the Laplacian $L$ instead of the coefficients $\theta_k$. Given original Laplacian $L$, features $X$ and parameters $\Gamma$, the function $\mathcal{F}(L, X, \Gamma)$ outputs the spectrum of updated Laplacian $\tilde{L}$, then the filter will be:
\begin{equation}
g_{\theta}(\Lambda) = \sum_{k=0}^{K-1} (\mathcal{F}(L, X, \Gamma))^k \label{eq:kthspectralkernel_new}.
\end{equation}
Finally, the SGC-LL layer is primarily formulated as:
\begin{equation}
Y = U g_{\theta}(\Lambda) U^T X =  U\sum_{k=0}^{K-1} (\mathcal{F}(L, X, \Gamma))^k U^T X. \label{eq:spectral_layer_new}
\end{equation}
Evaluating Eq.(\ref{eq:spectral_layer_new}) is of $\mathcal{O}(N^2)$ complexity due to the dense matrix multiplication $U^T X$. If $g_{\theta}(\tilde{L})$ was approximated as a polynomial function of $\tilde{L}$ that could be calculated recursively, the  complexity would be reduced to $\mathcal{O}(K)$  due to the sparsity of Laplacian $\tilde{L}$. We choose the same Chebychev expansion as \cite{defferrard2016convolutional} to compute polynomial $T_k(\tilde{L})X$ of order $k$.

\subsubsection{Training Metric for Graph Update} 
\label{subsec:metric_learning}

For graph structured data, the Euclidean distance is no longer a good metric to measure vertex similarity. Therefore, the distance metric need to be adaptive along with the task and the features during training.
In articles of metrics learning, the algorithms were divided into supervised and unsupervised learning \cite{wang2015survey}. The optimal metric obtained in unsupervised fashion minimizes the intra-cluster distances and also maximizes the inter-cluster distances. For labeled datasets, the learning objective is to find the metric minimizes the loss. \emph{Generalized Mahalanobis distance} between $x_i$ and $x_j$ is formulated as:
\begin{equation}
	\mathbb{D}(x_i, x_j) = \sqrt{(x_i-x_j)^T M (x_i-x_j)}. \label{eq:m_dist}
\end{equation}
If $M=I$, Eq.(\ref{eq:m_dist}) reduces to the Euclidean distance. In our model, the symmetric positive semi-definite matrix $M=W_d W_d^T$, where $W_d$ is one of the trainable weights of SGC-LL layer. The $W_d\in \mathbb{R}^{d\times d}$ is the transform basis to the space where we measure the Euclidean distance between $x_i$ and $x_j$. Then, we use the distance to calculate the Gaussian kernel: 
\begin{equation} 
\mathbb{G}_{x_i,x_j}= \exp(-\mathbb{D}(x_i, x_j)/(2\sigma^2)). \label{eq:G_m_dist}
\end{equation}
After normalization of $\mathbb{G}$, we obtain a dense adjacency matrix $\hat{A}$. In our model, the optimal metric $\hat{W}_d$ is the one that build the graph Laplacian set $\{\hat{L}\}$ minimizing the predictive losses. 

\subsubsection{Re-parameterization on feature transform} 
\label{subsec:feature_embedding}
In classical CNNs, the output feature of convolution layer is the sum of all the feature maps from last layer in which they were calculated by independent filters. This means the new features are not only built upon the neighbor vertices, but also depend on other intra-vertex features. However, on graph convolution, it is not explainable to create and train separate topological structures for different vertex features on the same graph. In order to construct mappings of both intra- and inter-vertex features, at SGC-LL layer, we introduce a transform matrix and bias vector applied on output features. Based on Eq.(\ref{eq:spectral_layer_new}), the re-parameterization on output feature is formulated as:
\begin{equation}
Y = \big(U g_{\theta}(\Lambda) U^T X\big)W + b. \label{eq:kthspectral_rp}
\end{equation}
 At $i$-th layer the transform matrix $W_i \in \mathbb{R}^{d_{i-1}\times d_{i}}$ and the bias $b_i \in \mathbb{R}^{d_i\times 1}$ are trained along with metrics $M_i$, where $d_i$ is the feature dimensionality. Totally, at each SGC-LL layer, we have the parameters $\{M_i, W_i, b_i\}$ of $\mathcal{O}(d_i d_{i-1})$ learning complexity, independent of input graph size or degree. At next SGC-LL layer, the spectral filter will be built in another feature domain with different metrics.

\begin{figure*}[ht!]
	\centering
	\includegraphics[width=0.55\textheight]{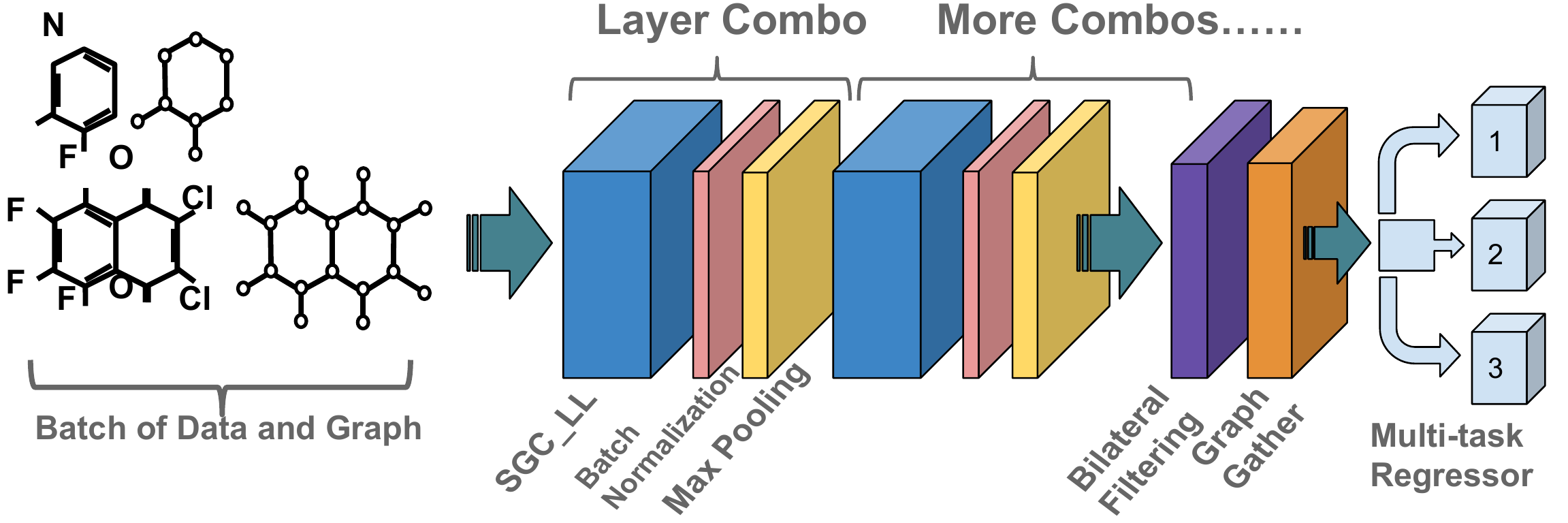}
	\caption{AGCN network configuration. Directly feed it on the original graph-structured data and their initial graphs. } \label{fig:AGCN_configuration}
\end{figure*}

\subsubsection{Residual Graph Laplacian} 
\label{subsec:residual_graph}

Some graph data have intrinsic graph structures, such as molecules.  Molecule is modeled as molecular graph with atom as vertex and bond as edge. Those bonds could be justified by chemical experiments. But, the most of data do not naturally have graph structure, so we have to construct graphs before feed them to the network. Besides above two cases, the most likely case is that the graphs created in unsupervised way can not sufficiently express all of the meaningful topological structure for specific task. Use chemical compound as example, the intrinsic graph given by SMILES \cite{weininger1988smiles} sequence does NOT tell anything about the toxicity of compound. Merely on intrinsic graph, it is hard to learn the meaningful representations of toxicity.

Because there is no prior knowledge on distance metric, the metrics $M$ are randomly initialized, so it may take long to converge. In order to accelerate the training and increase the stability of learned graph topology structure, we announce a reasonable assumption that the optimal graph Laplacian $\hat{L}$ is a small shifting from the original graph Laplacian $L$:
\begin{equation}
\hat{L}  = L + \alpha L_{res}
\end{equation}
In other words, the original graph Laplacian $L$ has disclosed a large amount of helpful graph structural information, except for those sub-structures consist of virtual vertex connections that can not be directly learned on intrinsic graph. Therefore, instead of learning $\hat{L}$, we learn the residual graph Laplacian $L_{res}(i) = \mathcal{L}(M_i, X)$, which is evaluated by Eq.(\ref{eq:G_m_dist}) and Eq.(\ref{eq:laplacian}). The influence of $L_{res}(i)$ on final graph topology is controlled by $\alpha$. 
The operations in SGC-LL layer are summarized as Algorithm \ref{alg:SGC_LL}.

\begin{algorithm}
	\caption{SGC-LL Layer}\label{alg:SGC_LL}
	\hspace*{\algorithmicindent} \textbf{Data} $\mathbf{X}=\{X_i\}, \mathbf{L}=\{L_i\}$,
\textbf{Parameter} $\alpha, M, W, b$
	\begin{algorithmic}[1]
		\For{$i$-th graph sample $X_i$ in mini-batch}
		\State $\tilde{A}_i \gets Eq.(\ref{eq:m_dist}), Eq.(\ref{eq:G_m_dist}) $ 
		\State $L_{res}(i) \gets I - \tilde{D}_i^{-1/2} \tilde{A}_i \tilde{D}_i^{-1/2} $ \Comment{$\tilde{D}_i=diag(\tilde{A}_i)$}
		\State $\tilde{L}_i = L_i + \alpha L_{res}(i)$ 
		\State $Y_i \gets Eq.(\ref{eq:kthspectral_rp})$
		\EndFor
		\State \textbf{return} $\mathbf{Y}=\{Y_i\}$
	\end{algorithmic}
\end{algorithm}

\subsection{AGCN Network } 
The proposed network is named as the Adaptive Graph Convolution Network (AGCN), because the SGC-LL layer is able to efficiently learn \emph{adaptive} graph topology structure  according to the data and the context of learning task. Besides SGC-LL layer, the AGCN has graph max pooling layer and graph gather layer \cite{gomes2017atomic}. 

\subsubsection{Graph Max Pooling} The graph max pooling is conducted feature-wise. For feature $x_v$ at $v$-th vertex of graph, the pooling replaces the $j$-th feature $x_v(j)$ with the maximum one among the $j$-th feature from its neighbor vertices and himself. If $N(v)$ is the set of neighbor vertices of $v$, the new feature at vertex $v$: $\hat{x}_v(j) = \max(\{x_v(j),  x_i(j) , \forall i \in N(v)\})$. 

\subsubsection{Graph Gather} The graph gather layer element-wise sums up all the vertex feature vectors as the representation of graph data. The output vector of gather layer will be used for graph-level prediction. Without the graph gather layer, the AGCN can also be trained and used for vertex-wise prediction tasks, given labels on vertex. The vertex-wise predictions include graph completion and many predictions on social networks. 

\subsubsection{Bilateral Filter}  The purpose of using bilateral filter layer \cite{gadde2016superpixel} in AGCN is to prevent over-fitting. The residual graph Laplacian definitely adapts the model to better fit the training task, but, at the risk of over-fitting.  To mitigate over-fitting, we introduce a revised bilateral filtering layer to regularize activation of SGC-LL layer by augmenting the spatial locality of $L$. We also introduced batch normalization layers to accelerate the training \cite{ioffe2015batch}.

\subsubsection{Network Configuration} The AGCN consists of multiple consecutive layer combos, the core layer of which is SGC-LL layer. The layer combo comprises one SGC-LL layer, one batch normalization layer \cite{ioffe2015batch} and one graph max pooling layer. See Figure. \ref{fig:AGCN_configuration} for illustration. A residual graph Laplacian is trained at each SGC-LL layer. At the graph pooling layer that follows, the adaptive graph (intrinsic + residual graph) is reused until next SGC-LL layer, because SGC-LL transform features, so the next SGC-LL need to retrain a new residual graph.

After passing a layer combo, the graph structures in batch will be updated, while the graph sizes remain. 
Because for data like organic compound, small sub-structures are decisive on specific chemical property, e.g toxicity. For instance, aromatic hydrocarbons are usually strongly toxic, while if the hydrogen (H) atom was replaced by methyl group (-CH3), their toxicity would be greatly reduced. Therefore, any graph coarsening or feature averaging will damage the completeness of those informative local structures. So, we choose max pooling and do not skip any vertex in convolution. In the paper, we test the network on graph-wise prediction tasks. So, the graph gather layer is the last layer before regressors.

\subsection{Batch Training of Diverse Graphs}
One of the greatest challenges for conducting convolution on graph-structured data is the difficulty of matching the diverse local topological structures of training samples: 1) bring extra difficulty of designing convolution kernel, because the invariance of kernel is not satisfied on graph, and the node indexing sometimes matters; 2) Resizing or reshaping of graph is not reasonable for some data e.g  molecules. Different from images and videos, which work with classical convolution on tensor, the compatibility with diverse topology is necessary for convolution on graph. The proposed SGC-LL layer train separate graph Laplacian, that preserve all local topological structures of data. Because we find that it is the feature space and the distance metrics that actually matter in constructing graph structure, the SGC-LL layer only requires all samples in batch to share the same feature transform matrix and distance metrics. Furthermore, the training parameter number is only dependent on feature dimensionality. Therefore, the AGCN accepts training batch consist of raw graph-structured data samples with different topology and size. It is noted that additional RAM consumption will be brought by the initial graph Laplacians that need to be constructed before training, and we still need to keep them for updating kernels. But, it is acceptable because graph Laplacians are usually sparse.

\section{Experiments}
\label{sec:exp}

In the experiments, we compared our AGCN network with the state-of-the-art graph CNNs.  \cite{bruna2013spectral} implemented convolution with a spectral filter formed by linear B-spline interpolation, referred as graphconv. Neural fingerprint \cite{duvenaud2015convolutional}, referred as NFP,  is the cutting-edge neural network for molecules. It uses kernel constructed in spatial domain.  We refer to the graph CNN equipped with $K$-localized spectral filter as GCN \cite{defferrard2016convolutional}. In this section, extensive numerical results show that our AGCN outperforms all existing graph CNNs, and we explain how the proposed SGC-LL layer boost the performance.

\begin{figure}[h!]
	\centering
	\includegraphics[width=0.45\textwidth]{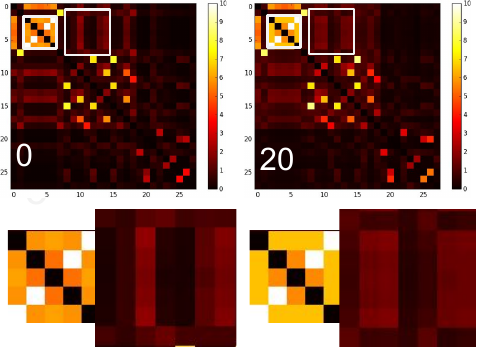}
	\caption{ Two heat maps of $28\times28$ similarity matrix $S$ of nodes of C20N2O5S (Benfuracarb). One (labeled by 0) is before training, and the other is after the first 20 epoch. Enlarged part of matrix better indicates the learning of graph.} \label{fig:heatmap}
\end{figure}

\subsection{Performance boosted by SGC-LL Layer}
\label{sec:LL}
\begin{figure}[h!]
	\centering
	\includegraphics[width=0.45\textwidth]{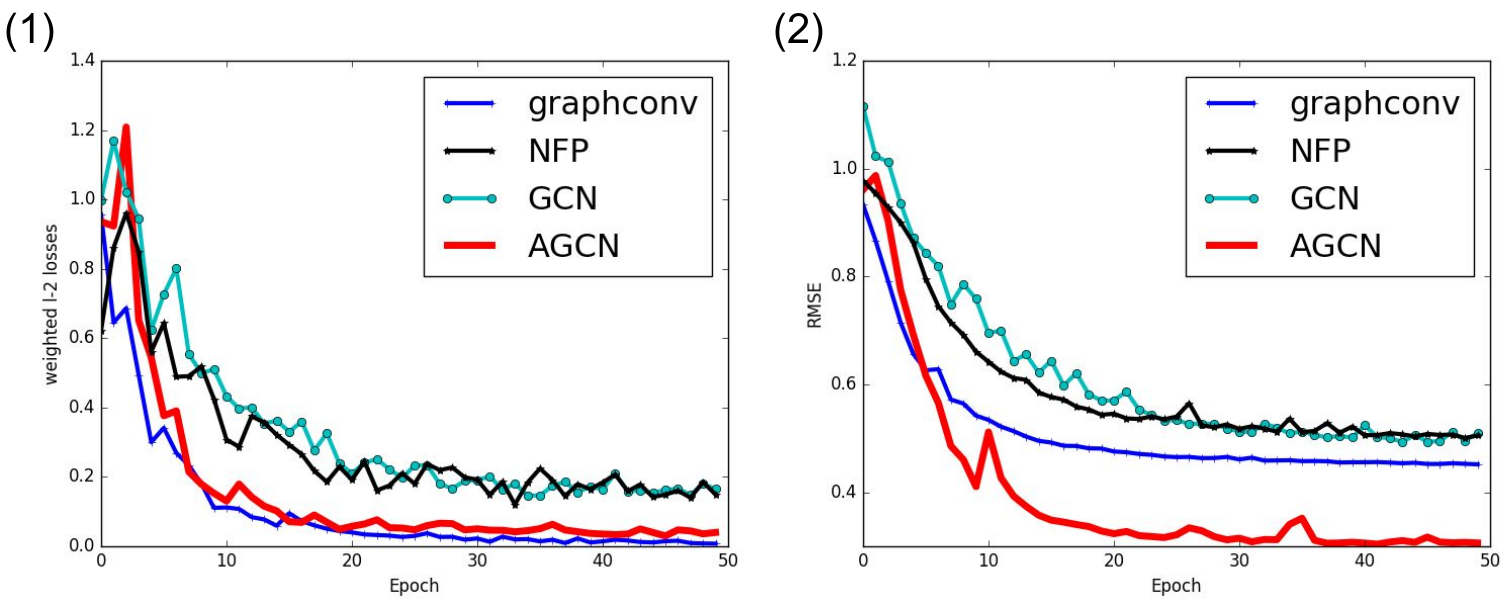}
	\caption{(1) training losses; (2) RMSE scores for solubility value prediction on delaney dataset \protect\cite{delaney2004esol}.} \label{fig:curves}
\end{figure}

The spectral filter at SGC-LL Layer is constructed on adaptive graph that consists of individual graph and residual graph. Individual graph is either intrinsic graph directly given by data itself or from clustering. Individual graph which enables the network to read data of different structure. Furthermore, the graphs will be updated during training since the network is trained to optimize the distance metric and feature transform for training data. The experiment demonstrated a close correlation between the updated graph  and network performance. In Figure. \ref{fig:heatmap}, if zoom in, it is easy to find the significant difference on node similarity after 20 epochs. This means the graph structure of compound in the trained distance metric has been updated.  In the meanwhile, the weighted $l_2$ losses dropped dramatically during the first 20 epoch, so did the mean RMSE score. Besides, the RMSE and losses curves proved that the AGCN (red line) has overwhelmingly better performance than other graph CNNs in both convergence speed and predictive accuracy (Figure. \ref{fig:curves}). We attribute this improvement to the \textbf{adaptive graph} and the learning of \textbf{residual Laplacian} at SGC-LL layer.

\begin{table*}[!ht]
	\centering
	\begin{tabular}{l|c|c|c| c }
		\hline 
		Datasets    & Delaney solubility & Az-logD & NCI & Hydration-free energy\\
		\hline \hline
		graphconv    &  0.4222 $\pm$ 8.38$\mathrm{e}{-2}$  &  0.7516 $\pm$ 8.42$\mathrm{e}{-3}$ & 0.8695 $\pm$ 3.55$\mathrm{e}{-3}$ & 2.0329 $\pm$ 2.70$\mathrm{e}{-2}$  \\ 
		NFP    &  0.4955 $\pm$ 2.30$\mathrm{e}{-3}$ & 0.9597 $\pm$ 5.70$\mathrm{e}{-3}$ & 0.8748 $\pm$ 7.50$\mathrm{e}{-3}$ & 3.4082 $\pm$ 3.95$\mathrm{e}{-2}$ \\ 
		GCN  &   0.4665 $\pm$ 2.07$\mathrm{e}{-3}$  & 1.0459 $\pm$ 3.92$\mathrm{e}{-3}$ & 0.8717 $\pm$ 4.14$\mathrm{e}{-3}$ & 2.2868 $\pm$ 1.37 $\mathrm{e}{-2}$ \\  
		AGCN      & \textbf{0.3061} $\pm$ 5.34$\mathrm{e}{-3}$ & \textbf{0.7362} $\pm$ 3.54$\mathrm{e}{-3}$ &  \textbf{0.8647} $\pm$ 4.67$\mathrm{e}{-3}$ & \textbf{1.3317} $\pm$ 2.73$\mathrm{e}{-2}$  \\
		\hline
	\end{tabular}
	\caption{Mean and standard deviation of RMSE on Delaney, Az-logD, NIH-NCI and Hydration-free energy Datasets. Compare AGCN with graphconv \protect\cite{bruna2013spectral}, NFP \protect\cite{duvenaud2015convolutional}, GCN  \protect\cite{defferrard2016convolutional} \label{tab:regression-table} }
\end{table*}
\begin{table*}[!ht]
	\centering
	\begin{tabular}{l|c|c|c|c|c|c|c|c}
		\hline
		Datasets    &\multicolumn{2}{|c|}{Tox21} & \multicolumn{2}{|c}{ClinTox} & \multicolumn{2}{|c}{Sider} & \multicolumn{2}{|c}{Toxcast}\\
		\hline
		&  Validation   & Testing  & Validation & Testing  & Validation & Testing & Validation & Testing  \\
		\hline \hline 
		graphconv & 0.7105  & 0.7023 &  0.7896 & 0.7069 & 0.5806 &  0.5642 & 0.6497 & 0.6496\\ 
		NFP  & 0.7502  & 0.7341 & 0.7356 & 0.7469 &  0.6049 & 0.5525 & 0.6561 & 0.6384  \\
		GCN    &  0.7540  & 0.7481 & 0.8303 & 0.7573 & 0.6085 & 0.5914 & 0.6914 & 0.6739 \\  
		AGCN     &  \textbf{0.7947}    & \textbf{0.8016} &  \textbf{0.9267}  & \textbf{0.8678} & \textbf{0.6112} & \textbf{0.5921} & \textbf{0.7227} & \textbf{0.7033}  \\
		\hline
	\end{tabular}
	\caption{Task-averaged ROC-AUC Scores on Tox21, ClinTox, Sider, Toxcast Datasets . The same benchmarks as Table. \ref{tab:regression-table}.  	\label{tab:classification}}
\end{table*}

\subsection{Multi-task Prediction on Molecular Datasets}
Delaney Dataset \cite{delaney2004esol} contains aequeous solubility data for 1,144 low molecular weight compounds. The largest compound in the dataset has 492 atoms, while the smallest only consists of 3 atoms. NCI Database has around 20,000 compounds and 60 prediction tasks from drug reaction experiments to clinical pharmacology studies. At last, Az-logD dataset from ADME \cite{vugmeyster2012absorption} offers the logD measurements on permeability for 4200 compounds. Besides, we also have a small dataset of 642 compounds for hydration-free energy study. The presented task-averaged RMSE scores and standard deviations were obtained after 5-fold cross-validation.

Tox21 Dataset \cite{mayr2016deeptox} contains 7,950 chemical compounds and labels for classifications on 12 essays of toxicity. However, additional difficulty comes from the missing labels for part of the 12 tasks. For those with missing labels, we excluded them from loss computation, but still kept them in training set. ClinTox is a public dataset of 1451 chemical compounds for clinical toxicological study together with labels for 2 tasks. Sider \cite{kuhn2010side} database records 1392 drugs and their 27 different side effects or adverse reactions. Toxcast \cite{dix2006toxcast} is another toxicological research database that has 8,599 SMILES together with labels for 617 predictive tasks. For $N$-task prediction, the network graph model will become an analog of K-ary tree with $N$ leaf nodes, each of which comprises a fully connected layer and a logistic regression for each task.  

To prove the advantages of AGCN, we compared it with three state-of-the-art graph CNN benchmarks: the first spectral graph CNN (graphconv) with spline interpolated kernel \cite{bruna2013spectral}, the extension to $k$-localized spectral filter (GCN) \cite{defferrard2016convolutional} and neural fingerprint (NFP) \cite{duvenaud2015convolutional}, the cutting-edge neural network for molecules. In Table. \ref{tab:regression-table}, our AGCN reduced the mean RMSE by 31$\%$ -40$\%$ on Delaney dataset, averagely 15$\%$ on Az-logD and 2$\sim$4$\%$ on testing set of NCI. 
It looks the adaptive graph and the residual Laplacian learning for hidden structures are more useful when data is short.
As to the multi-task classification results from Table. \ref{tab:classification}, we notice that the AGCN significantly boosted the accuracy on both small and large datasets. For the mass of 617 tasks of Toxcast, the performance of classifier still got improved by 3$\%$ (0.03) on average, compared to the state-of-the-arts.

Molecular graph, directly given by chemical formula, is the intrinsic graph for compound data. They come with high variety in both topological structure and graph size. The spectral kernel in \emph{graphconv} \cite{bruna2013spectral} can only connect 1-hop neighbor (nearby vertex directly connected by edge), so it is over-localized. This becomes an issue when dealing with molecules, because some important sub-structures of molecular graph are impossible to be covered by over-localized kernels. For example, centered at any carbon atom of Benzene ring (C6H6), the kernel at least needs to cover the vertices of distance $d_{\mathcal{G}}<=$3, if you want to learn representation from the ring as a whole. The $K$-localized kernel in GCN \cite{defferrard2016convolutional} is no longer too local, but the kernel is still assumed to be shared among data. It is fine if the molecules in training set share many common sub-structures such as OH (carbonyl group) and C6H6 (Benzene). See Figure. \ref{fig:kernels} for illustration. However, if the molecules are from different classes of compound, GCN may not work well especially when data from some type are short. This is probably why the GCN has similar performance as AGCN on large datasets such as the Sider, but it dramatically worsened on small datasets, e.g Delaney and Clintox. 

The AGCN is able to handle molecular data in a better way. The adaptive graph allows input samples to have unique graph Laplacian, so each compound indeed has its unique convolution filter customized according to its unique topological structure. Because of this capability, we can feed the network on the original data (atom/edge features and molecular graph) without any loss of information.
Furthermore, our SGC-LL layers train the distance metric minimizing predictive losses of specific tasks together with other transform parameters. Therefore, when it converged, at each SGC-LL, we would find the optimal feature space and distance metric to construct the graph that best serve the task, e.g. toxicity and solubility prediction. This learned graph may contain new edges that did not exist in original molecular graph.

\subsection{Point Cloud Object Classification}
\begin{figure}[h!]
	\centering
	\includegraphics[width=0.33\textheight]{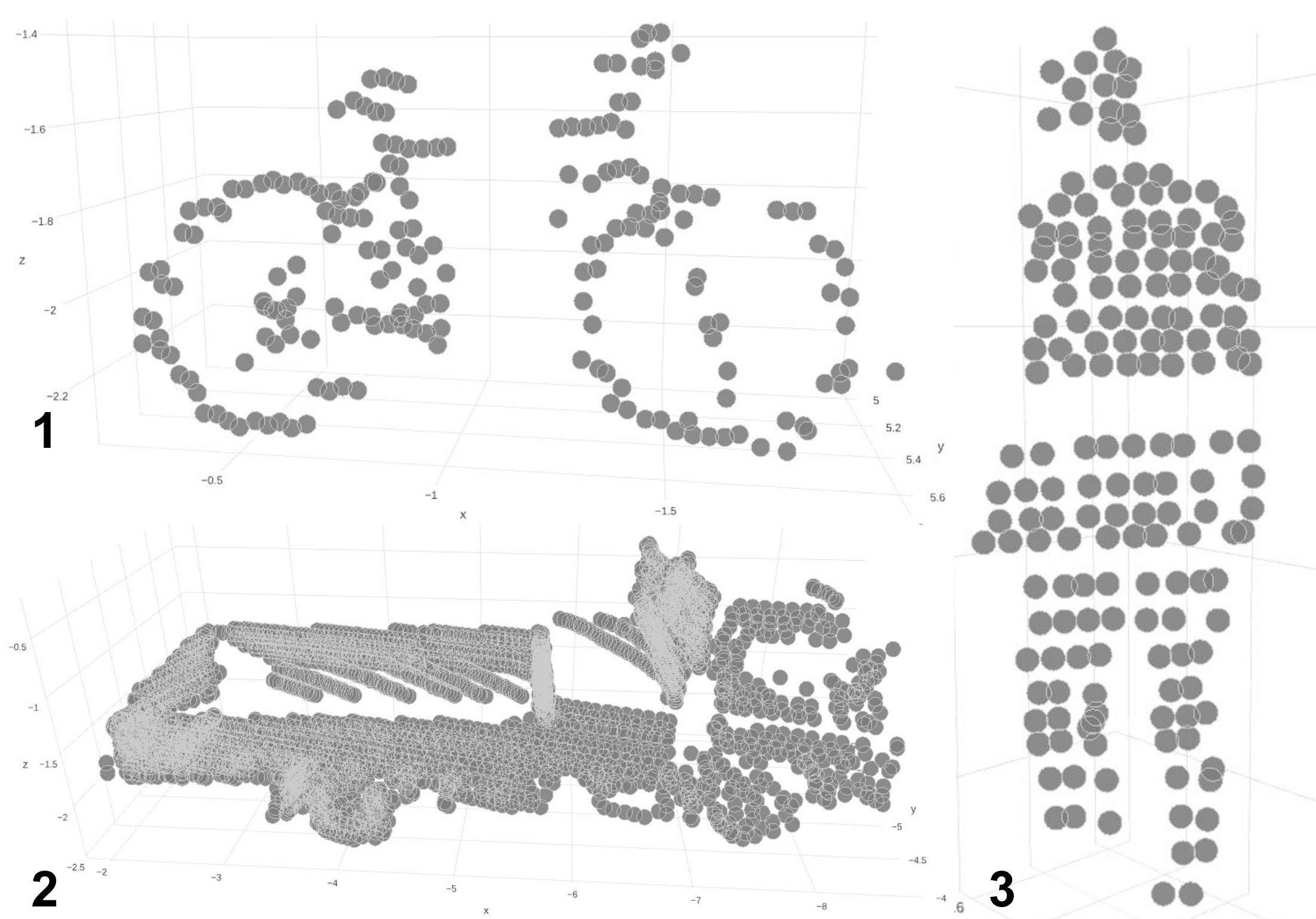}
	\caption{Point cloud examples of different size. 1: bicycle, 124 points; 2: truck, 615 points; 3: pedestrian, 78 points.} \label{fig:example_pcd}
\end{figure}

The Sydney urban point cloud dataset contains street objects scanned with a Velodyne HDL-64E LIDAR, collected in the CBD of Sydney, Australia. There are 631 individual scans of objects across 26 classes. Due to the actual size and shape of object highly differ, the numbers of received point for different objects also vary (see Figure. \ref{fig:example_pcd} for illustration). 

Before feed point sets to previous CNNs, we need to unify the size by downsampling. Coarsened samples must lose part of structural information. While, the AGCN overcomes such drawback by accepting raw point sets of different size. Previous graph convolution share an identical kernel, but, the shared one may mix up features on points regardless of the actual distance. While, the AGCN is able to do convolution exactly according to the spatial relations. The initial graphs of point cloud were constructed by agglomerative clustering. The cutting-edge method on point set recognition, PointNet \cite{qi2016pointnet}, cannot handle varying sized point cloud data.
\begin{table}[h!]
	\centering
	\begin{tabular}{c|c|c|c}
		\hline
		   & All Classes & Building & Traffic Light\\
		\hline 
		graphconv    & 0.6523 & 0.6754 & 0.5197\\ 
		NFP   & 0.6350 & 0.8037 & 0.5344 \\
		GCN      & 0.6657 &  0.8427 & 0.7417 \\  
		AGCN       & \textbf{0.6942} &  \textbf{0.9906} & \textbf{0.8556} \\
		\hline
	\end{tabular}
		\caption{Average ROC-AUC Scores on testing set of Sydney Urban Objects Dataset.The same benchmarks as Table. \ref{tab:regression-table}.} \label{table:pcd_sdney}
\end{table}

After 5-fold cross-validation, averaged ROC-AUC scores were calculated on a testing set that has 200 samples. From Table. \ref{table:pcd_sdney}, we see the AGCN outperformed other graph CNNs by $3\sim 6\%$ on all classes average score. For specific large objects like building, we have the AUC score close to 1, while other networks did worse because they have to coarsen the graphs first. For important road objects such as traffic light, the AGCN also lifted the classification accuracy by at least $10\%$ in terms of ROC-AUC. It sufficiently showed that the AGCN can extract more meaningful features than other graph CNNs on point clouds. The information completeness of data fed to the AGCN also benefit the performance, which is attributed to the adaptive graphs constructed and learned at proposed SGC-LL layers.

\section{Acknowledgments}
This work was partially supported by US National Science Foundation IIS-1423056, CMMI-1434401, CNS-1405985, IIS-1718853 and the NSF CAREER grant IIS-1553687.

\section{Conclusions}
We proposed a novel spectral graph convolver (SGC-LL) that work with adaptive graphs. SGC-LL learns the residual graph Laplacian via learning the optimal metric and feature transform.  As far as we know, the AGCN is the first graph CNN that accepts data of arbitrary graph structure and size. The supervised training of residual Laplacian drives the model to better fit the prediction task. The extensive multi-task learning experiments on various graph-structured data indicated that the AGCN outperformed the state-of-the-art graph CNN models on various prediction tasks.

\bibliographystyle{aaai}
\bibliography{ref}

\begin{thebibliography}{}

\bibitem[\protect\citeauthoryear{Altae-Tran \bgroup et al\mbox.\egroup
  }{2016}]{altae2016low}
Altae-Tran, H.; Ramsundar, B.; Pappu, A.~S.; and Pande, V.
\newblock 2016.
\newblock Low data drug discovery with one-shot learning.
\newblock {\em arXiv preprint arXiv:1611.03199}.

\bibitem[\protect\citeauthoryear{Bruna \bgroup et al\mbox.\egroup
  }{2013}]{bruna2013spectral}
Bruna, J.; Zaremba, W.; Szlam, A.; and LeCun, Y.
\newblock 2013.
\newblock Spectral networks and locally connected networks on graphs.
\newblock {\em arXiv preprint arXiv:1312.6203}.

\bibitem[\protect\citeauthoryear{Chung}{1997}]{chung1997spectral}
Chung, F.~R.
\newblock 1997.
\newblock {\em Spectral graph theory}.
\newblock Number~92. American Mathematical Soc.

\bibitem[\protect\citeauthoryear{De~Brabandere \bgroup et al\mbox.\egroup
  }{2016}]{de2016dynamic}
De~Brabandere, B.; Jia, X.; Tuytelaars, T.; and Van~Gool, L.
\newblock 2016.
\newblock Dynamic filter networks.
\newblock In {\em Neural Information Processing Systems (NIPS)}.

\bibitem[\protect\citeauthoryear{Defferrard, Bresson, and
  Vandergheynst}{2016}]{defferrard2016convolutional}
Defferrard, M.; Bresson, X.; and Vandergheynst, P.
\newblock 2016.
\newblock Convolutional neural networks on graphs with fast localized spectral
  filtering.
\newblock In {\em Advances in Neural Information Processing Systems},
  3837--3845.

\bibitem[\protect\citeauthoryear{Delaney}{2004}]{delaney2004esol}
Delaney, J.~S.
\newblock 2004.
\newblock Esol: estimating aqueous solubility directly from molecular
  structure.
\newblock {\em Journal of chemical information and computer sciences}
  44(3):1000--1005.

\bibitem[\protect\citeauthoryear{Dix \bgroup et al\mbox.\egroup
  }{2006}]{dix2006toxcast}
Dix, D.~J.; Houck, K.~A.; Martin, M.~T.; Richard, A.~M.; Setzer, R.~W.; and
  Kavlock, R.~J.
\newblock 2006.
\newblock The toxcast program for prioritizing toxicity testing of
  environmental chemicals.
\newblock {\em Toxicological Sciences} 95(1):5--12.

\bibitem[\protect\citeauthoryear{Dundar \bgroup et al\mbox.\egroup
  }{2015}]{dundar2015simplicity}
Dundar, M.; Kou, Q.; Zhang, B.; He, Y.; and Rajwa, B.
\newblock 2015.
\newblock Simplicity of kmeans versus deepness of deep learning: A case of
  unsupervised feature learning with limited data.
\newblock In {\em Machine Learning and Applications (ICMLA), 2015 IEEE 14th
  International Conference on},  883--888.
\newblock IEEE.

\bibitem[\protect\citeauthoryear{Duvenaud \bgroup et al\mbox.\egroup
  }{2015}]{duvenaud2015convolutional}
Duvenaud, D.~K.; Maclaurin, D.; Iparraguirre, J.; Bombarell, R.; Hirzel, T.;
  Aspuru-Guzik, A.; and Adams, R.~P.
\newblock 2015.
\newblock Convolutional networks on graphs for learning molecular fingerprints.
\newblock In {\em Advances in neural information processing systems},
  2224--2232.

\bibitem[\protect\citeauthoryear{Gadde \bgroup et al\mbox.\egroup
  }{2016}]{gadde2016superpixel}
Gadde, R.; Jampani, V.; Kiefel, M.; Kappler, D.; and Gehler, P.~V.
\newblock 2016.
\newblock Superpixel convolutional networks using bilateral inceptions.
\newblock In {\em European Conference on Computer Vision},  597--613.
\newblock Springer.

\bibitem[\protect\citeauthoryear{Gomes \bgroup et al\mbox.\egroup
  }{2017}]{gomes2017atomic}
Gomes, J.; Ramsundar, B.; Feinberg, E.~N.; and Pande, V.~S.
\newblock 2017.
\newblock Atomic convolutional networks for predicting protein-ligand binding
  affinity.
\newblock {\em arXiv preprint arXiv:1703.10603}.

\bibitem[\protect\citeauthoryear{Henaff, Bruna, and
  LeCun}{2015}]{henaff2015deep}
Henaff, M.; Bruna, J.; and LeCun, Y.
\newblock 2015.
\newblock Deep convolutional networks on graph-structured data.
\newblock {\em arXiv preprint arXiv:1506.05163}.

\bibitem[\protect\citeauthoryear{Hinton \bgroup et al\mbox.\egroup
  }{2012}]{hinton2012deep}
Hinton, G.; Deng, L.; Yu, D.; Dahl, G.~E.; Mohamed, A.-r.; Jaitly, N.; Senior,
  A.; Vanhoucke, V.; Nguyen, P.; Sainath, T.~N.; et~al.
\newblock 2012.
\newblock Deep neural networks for acoustic modeling in speech recognition: The
  shared views of four research groups.
\newblock {\em IEEE Signal Processing Magazine} 29(6):82--97.

\bibitem[\protect\citeauthoryear{Ioffe and Szegedy}{2015}]{ioffe2015batch}
Ioffe, S., and Szegedy, C.
\newblock 2015.
\newblock Batch normalization: Accelerating deep network training by reducing
  internal covariate shift.
\newblock {\em arXiv preprint arXiv:1502.03167}.

\bibitem[\protect\citeauthoryear{Kipf and Welling}{2016}]{kipf2016semi}
Kipf, T.~N., and Welling, M.
\newblock 2016.
\newblock Semi-supervised classification with graph convolutional networks.
\newblock {\em arXiv preprint arXiv:1609.02907}.

\bibitem[\protect\citeauthoryear{Kuhn \bgroup et al\mbox.\egroup
  }{2010}]{kuhn2010side}
Kuhn, M.; Campillos, M.; Letunic, I.; Jensen, L.~J.; and Bork, P.
\newblock 2010.
\newblock A side effect resource to capture phenotypic effects of drugs.
\newblock {\em Molecular systems biology} 6(1):343.

\bibitem[\protect\citeauthoryear{Looks \bgroup et al\mbox.\egroup
  }{2017}]{looks2017deep}
Looks, M.; Herreshoff, M.; Hutchins, D.; and Norvig, P.
\newblock 2017.
\newblock Deep learning with dynamic computation graphs.
\newblock {\em arXiv preprint arXiv:1702.02181}.

\bibitem[\protect\citeauthoryear{Mayr \bgroup et al\mbox.\egroup
  }{2016}]{mayr2016deeptox}
Mayr, A.; Klambauer, G.; Unterthiner, T.; and Hochreiter, S.
\newblock 2016.
\newblock Deeptox: toxicity prediction using deep learning.
\newblock {\em Frontiers in Environmental Science} 3:80.

\bibitem[\protect\citeauthoryear{Qi \bgroup et al\mbox.\egroup
  }{2016}]{qi2016pointnet}
Qi, C.~R.; Su, H.; Mo, K.; and Guibas, L.~J.
\newblock 2016.
\newblock Pointnet: Deep learning on point sets for 3d classification and
  segmentation.
\newblock {\em arXiv preprint arXiv:1612.00593}.

\bibitem[\protect\citeauthoryear{Shuman \bgroup et al\mbox.\egroup
  }{2013}]{shuman2013emerging}
Shuman, D.~I.; Narang, S.~K.; Frossard, P.; Ortega, A.; and Vandergheynst, P.
\newblock 2013.
\newblock The emerging field of signal processing on graphs: Extending
  high-dimensional data analysis to networks and other irregular domains.
\newblock {\em IEEE Signal Processing Magazine} 30(3):83--98.

\bibitem[\protect\citeauthoryear{Simonovsky and
  Komodakis}{2017}]{simonovsky2017dynamic}
Simonovsky, M., and Komodakis, N.
\newblock 2017.
\newblock Dynamic edge-conditioned filters in convolutional neural networks on
  graphs.
\newblock {\em arXiv preprint arXiv:1704.02901}.

\bibitem[\protect\citeauthoryear{Vugmeyster, Harrold, and
  Xu}{2012}]{vugmeyster2012absorption}
Vugmeyster, Y.; Harrold, J.; and Xu, X.
\newblock 2012.
\newblock Absorption, distribution, metabolism, and excretion (adme) studies of
  biotherapeutics for autoimmune and inflammatory conditions.
\newblock {\em The AAPS journal} 14(4):714--727.

\bibitem[\protect\citeauthoryear{Wallach, Dzamba, and
  Heifets}{2015}]{wallach2015atomnet}
Wallach, I.; Dzamba, M.; and Heifets, A.
\newblock 2015.
\newblock Atomnet: a deep convolutional neural network for bioactivity
  prediction in structure-based drug discovery.
\newblock {\em arXiv preprint arXiv:1510.02855}.

\bibitem[\protect\citeauthoryear{Wang and Sun}{2015}]{wang2015survey}
Wang, F., and Sun, J.
\newblock 2015.
\newblock Survey on distance metric learning and dimensionality reduction in
  data mining.
\newblock {\em Data Mining and Knowledge Discovery} 29(2):534--564.

\bibitem[\protect\citeauthoryear{Weininger}{1988}]{weininger1988smiles}
Weininger, D.
\newblock 1988.
\newblock Smiles, a chemical language and information system. 1. introduction
  to methodology and encoding rules.
\newblock {\em Journal of chemical information and computer sciences}
  28(1):31--36.

\bibitem[\protect\citeauthoryear{Weiss, Torralba, and
  Fergus}{2009}]{weiss2009spectral}
Weiss, Y.; Torralba, A.; and Fergus, R.
\newblock 2009.
\newblock Spectral hashing.
\newblock In {\em Advances in neural information processing systems},
  1753--1760.

\bibitem[\protect\citeauthoryear{Wu \bgroup et al\mbox.\egroup
  }{2017}]{wu2017moleculenet}
Wu, Z.; Ramsundar, B.; Feinberg, E.~N.; Gomes, J.; Geniesse, C.; Pappu, A.~S.;
  Leswing, K.; and Pande, V.
\newblock 2017.
\newblock Moleculenet: A benchmark for molecular machine learning.
\newblock {\em arXiv preprint arXiv:1703.00564}.

\end{thebibliography}

\end{document}